\documentclass{article}

\usepackage{arxiv}

\usepackage[utf8]{inputenc} % allow utf-8 input
\usepackage[T1]{fontenc}    % use 8-bit T1 fonts
\usepackage{hyperref}       % hyperlinks
\usepackage{url}            % simple URL typesetting
\usepackage{booktabs}       % professional-quality tables
\usepackage{amsfonts}       % blackboard math symbols
\usepackage{nicefrac}       % compact symbols for 1/2, etc.
\usepackage{microtype}      % microtypography
\usepackage{lipsum}
\usepackage{graphicx}
\graphicspath{ {./images/} }
\usepackage{graphicx}
\usepackage{amsmath}
\usepackage{booktabs}
\usepackage{microtype}
\usepackage{adjustbox}

\title{LLM-Assisted Behavioural and Scenario Augmentation for Agent-Based Energy Adoption Models}

\author{
Iias Faiud$^{1}$ \quad
Hossein Khaleghy$^{1}$ \quad
Michael Schukat$^{1}$ \quad
Karl Mason$^{1}$
\\[1ex]
$^{1}$School of Computer Science, University of Galway \\
Galway, Ireland, H91 TK33 \\
Iias Faiud: \texttt{i.faiud1@universityofgalway.ie}
}

\begin{document}
\maketitle
\begin{abstract}
Recent advances in large language models (LLMs) create opportunities to enrich simulation-based energy policy analysis, particularly by supporting structured behavioural assumptions and exploratory techno-economic scenarios. However, directly replacing adoption models with LLM reasoning raises concerns regarding interpretability, reproducibility, and behavioural validity. This paper proposes a hybrid framework for LLM-assisted specification design, integrating bounded behavioural rubrics and structured scenario specifications into a calibrated agent-based model (ABM) of solar photovoltaic (PV) adoption by Irish dairy farms. The proposed approach preserves the original techno-economic adoption mechanism while augmenting it with bounded behavioural modulation and scenario-driven uncertainty analysis. Behavioural effects are represented through interpretable conservative, balanced, and optimistic rubrics, while future policy and market conditions are explored through fixed, rule-validated scenario specifications. Experimental results across multiple policy settings, Monte Carlo worlds, and random seeds demonstrate stable and economically plausible behaviour, with adoption outcomes remaining bounded and monotonic across behavioural regimes. The framework achieves up to approximately 13\% behavioural adoption increase relative to the corresponding logistic case without producing unstable or unrealistic saturation dynamics. The results demonstrate that LLM-assisted specifications can be integrated into calibrated energy ABMs in a controlled, reproducible, and policy-relevant manner.


\end{abstract}

% keywords can be removed

\section{Introduction}

Solar photovoltaic (PV) adoption is an important component of the transition toward low-carbon energy systems, and understanding adoption behaviour is central to effective energy policy design. In agricultural sectors such as dairy farming, PV decisions are shaped by techno-economic, financial, and behavioural factors, including electricity prices, capital costs, subsidies, financing conditions, and perceived investment risk. Agent-based models (ABMs) are widely used for studying innovation and energy technology diffusion because they can represent heterogeneous decision-makers, local interactions, and emergent adoption dynamics \cite{kiesling2012agent,rai2015agent,palmer2015modeling}.

Despite their strengths, many ABMs rely on fixed behavioural assumptions and manually specified scenario structures. These assumptions may limit the ability to explore changing behavioural responsiveness and diverse future policy-market conditions. Recent large language models (LLMs) offer new opportunities for generating structured behavioural descriptions, narratives, and contextual reasoning within simulation environments \cite{gao2024large,gurcan2024llm,park2023generative}. However, directly replacing calibrated behavioural or economic models with unconstrained LLM reasoning raises concerns regarding reproducibility, interpretability, calibration validity, and behavioural realism \cite{ji2023survey,larooij2025large,wang2025large}. These concerns are especially important in policy-oriented energy modelling, where transparency, stability, and uncertainty quantification are required.

This paper addresses this gap by proposing a hybrid framework that integrates LLM-assisted behavioural and scenario augmentation into a calibrated ABM of solar PV adoption by Irish dairy farms. Rather than replacing the underlying adoption mechanism, the framework preserves the calibrated techno-economic model and introduces bounded behavioural modulation through interpretable conservative, balanced, and optimistic rubrics. It also incorporates structured techno-economic scenarios representing alternative policy and market environments, including energy crises, financial tightening, subsidy withdrawal, green transition acceleration, and weak export incentives.

The framework is evaluated across multiple policy settings, behavioural regimes, Monte Carlo worlds, and random seeds using a reproducible uncertainty-aware simulation pipeline. Results show stable and monotonic behavioural ordering across all evaluated configurations, with behavioural effects remaining bounded and free from unrealistic saturation. Under favourable conditions, behavioural augmentation achieves adoption increases of approximately 13\% relative to the corresponding logistic case while preserving plausible adoption-cost relationships.

The main contributions of this work are as follows:
\begin{itemize}
    \item A controlled LLM-assisted specification-design workflow for constructing bounded behavioural rubrics and exploratory techno-economic scenarios for ABM-based policy simulation.
    \item A behavioural augmentation mechanism that preserves the calibrated logistic adoption model while enabling interpretable conservative, balanced, and optimistic responsiveness regimes.
    \item A robustness evaluation demonstrating stable behavioural ordering, bounded adoption dynamics, and reproducible scenario-driven policy analysis across multiple Monte Carlo worlds and random seeds.
\end{itemize}

\section{Related Work}
\label{sec:related_work}

Agent-based models have been widely used to study innovation diffusion and energy technology adoption. Building on diffusion-of-innovations theory, they allow heterogeneous decision-makers, social interactions, bounded rationality, and emergent aggregate adoption patterns to be represented explicitly \cite{rogers2014diffusion,kiesling2012agent}. In the PV adoption literature, ABMs have been applied to model adoption decisions using financial, behavioural, spatial, social, and environmental factors \cite{palmer2015modeling,robinson2015determinants,rai2015agent}. More recent studies have extended this line of work through calibrated policy evaluation, hybrid machine-learning and ABM architectures, and real-options-based adoption modelling \cite{pearce2018feed,peralta2022spatio,zhang2022agent}.

A related body of work emphasises the importance of empirical grounding, calibration, validation, and transparent documentation in ABM-based policy analysis. Prior studies argue that ABMs used for decision support require explicit calibration procedures, sensitivity analysis, and validation against observed data rather than relying only on plausible micro-level rules \cite{thiele2014facilitating,zhang2019empirically,grimm2020odd,troost2023keep}. This is particularly important in energy modelling, where scenario assumptions and uncertainty treatment can strongly affect policy conclusions \cite{pfenninger2014energy,yue2018review}. Scenario analysis is therefore commonly used to explore alternative futures, but such scenarios should be interpreted as structured exploratory assumptions rather than deterministic forecasts \cite{binsted2020evaluating}.

Recent work has begun to examine the role of large language models (LLMs) in simulation and agent-based modelling. Generative-agent studies show that LLMs can support richer behavioural representation in simulated environments \cite{park2023generative}, while broader surveys identify opportunities for LLM-assisted simulation across social, cyber, physical, and hybrid systems \cite{gao2024large}. Other studies explore LLMs as synthetic respondents or simulated social actors \cite{aher2023using,argyle2023out}. At the same time, the literature highlights concerns around hallucination, bias, demographic flattening, reproducibility, and weak validation when LLMs are used as direct substitutes for human behaviour or empirical models \cite{ji2023survey,wang2025large,larooij2025large}. These concerns suggest that LLMs should be used cautiously in policy-oriented simulation, particularly when model outputs may inform real-world decisions.

Taken together, existing work highlights a methodological tension. ABMs provide empirical grounding, heterogeneity, and uncertainty-aware policy evaluation, while LLMs offer new possibilities for structured behavioural and scenario specification. However, unconstrained LLM-based simulation can weaken reproducibility, interpretability, and validation. The approach proposed in this paper therefore uses LLM assistance only for bounded specification design: behavioural rubrics and structured scenario definitions. The calibrated ABM remains the core simulation engine, while the LLM-assisted components are converted into explicit, rule-validated, and reproducible inputs.

\section{Methodology}
\label{sec:methodology}

\begin{figure}[htbp]
\centering
\includegraphics[width=0.9\linewidth]{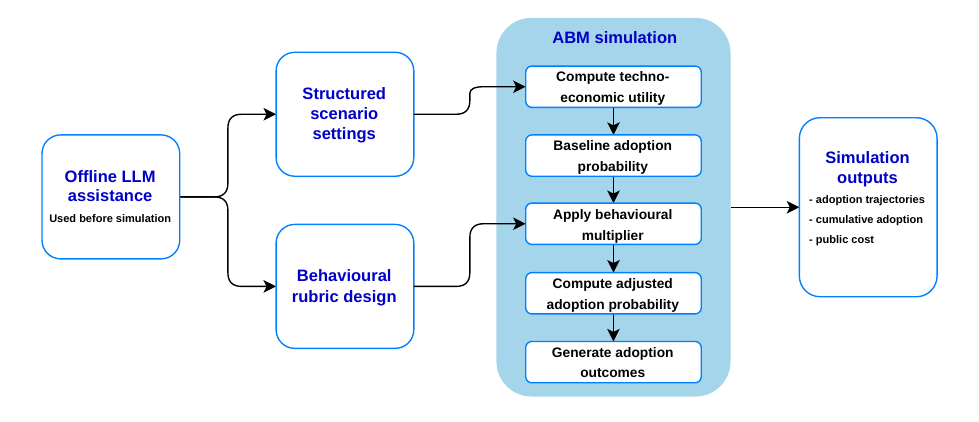}
\caption{Integration of offline LLM-assisted specification design with the calibrated ABM. Structured scenario settings modify existing techno-economic and policy inputs, while the behavioural rubric provides the bounded multiplier applied to the calibrated baseline adoption probability. The specifications are checked before simulation, and the LLM is not queried during ABM execution.}
\label{fig:LLM_diagram}
\end{figure}

This study builds on a calibrated ABM of PV adoption by Irish dairy farms~\cite{faiud2024agent,faiud2023modelling}. The proposed framework does not replace the calibrated adoption model. Instead, it augments it with bounded and interpretable behavioural specifications and structured scenario specifications. The framework consists of three components: a calibrated techno-economic adoption probability, a behavioural augmentation layer, and structured scenario specifications evaluated through Monte Carlo simulation. Figure~\ref{fig:LLM_diagram} illustrates how the offline LLM-assisted specifications are integrated into the ABM simulation workflow.

\subsection{Baseline Agent-Based Model}
\label{sec:baseline_abm}

The baseline model represents PV adoption as a probabilistic decision process driven by techno-economic utility. For each simulated adoption decision, the calibrated model computes a baseline adoption probability as:

\begin{equation}
p_{\text{base}} = \frac{1}{1 + \exp[-(\alpha U + \beta)]}
\label{eq:baseline_logistic}
\end{equation}
where \(U\) denotes the net-present-value-based economic utility of adopting PV, and \(\alpha\) and \(\beta\) are calibrated parameters from the baseline ABM. These parameters are held fixed throughout all experiments, ensuring that the LLM-assisted components act as augmentation layers rather than replacements for the empirical adoption mechanism.

The utility calculation reflects the investment conditions faced by dairy farms, including PV system costs, electricity price savings, grant support, feed-in-tariff (FiT) revenue, financing conditions, and operating costs. The model is evaluated up to 2040, with outcomes reported as cumulative adopters and cumulative public cost. Public cost includes both grant expenditure and FiT-related expenditure in the reported experiments.

\subsection{LLM Prompting and Specification Construction}
\label{sec:llm_prompting}

The LLM component is used as an offline specification-design aid rather than as an online simulation agent. The LLM is not queried during ABM execution and does not directly generate adoption decisions. Instead, it assists the construction of behavioural rubrics and techno-economic scenario specifications, which are manually inspected, fixed, and evaluated as deterministic model inputs.

The prompting process followed a constrained JSON-based workflow. For behavioural augmentation, the LLM was prompted to generate conservative, balanced, and optimistic rubrics using only predefined fields: sensitivity, multiplier bounds, and factor weights for payback attractiveness, grant support, export-tariff attractiveness, financing attractiveness, and electricity price pressure. The prompt required valid JSON, weights summing to one, no additional variables, and bounded adjustment of the calibrated logistic adoption probability.

For scenario specification, the final configurations were restricted to existing ABM inputs, including grant support, FiT level, loan rate, electricity price growth, PV cost trajectory, and export-related assumptions. Candidate configurations were reviewed for validity, interpretability, boundedness, plausibility, and consistency with the calibrated ABM. Invalid or implausible specifications were rejected or revised before being fixed as experimental inputs.

Thus, the contribution is not an autonomous LLM-agent framework, but a controlled integration pattern in which LLM assistance supports structured specification design within a calibrated and uncertainty-aware ABM. This ensures that stochastic variation arises from the ABM and Monte Carlo evaluation, not from live LLM generation. The prompting protocol is summarised in~\ref{app:llm_prompting}.

\subsection{LLM-Assisted Behavioural Augmentation}
\label{sec:behavioural_augmentation}

The behavioural augmentation layer applies the fixed rubric parameters as bounded multiplicative adjustments to the calibrated logistic adoption probability. Table~\ref{tab:behavioural_rubrics} reports the final rubric parameters used in the experiments. The weights in each rubric sum to one, and the multiplier bounds are designed to allow the behavioural layer to amplify or dampen adoption probabilities without overriding the calibrated logistic model.

\begin{table}[htbp]
\centering
\caption{Final behavioural rubric parameters used in the experiments. Min. and Max. denote the lower and upper bounds of the behavioural multiplier.}
\label{tab:behavioural_rubrics}
\setlength{\tabcolsep}{4pt}
\renewcommand{\arraystretch}{1.1}
\begin{adjustbox}{max width=\linewidth}
\begin{tabular}{lrrrrrrrr}
\toprule
Rubric & Sens. & Min. & Max. & Payback & Grant & FiT/export & Loan & Elec. price \\
\midrule
Conservative & 0.25 & 0.80 & 1.25 & 0.35 & 0.20 & 0.15 & 0.15 & 0.15 \\
Balanced     & 0.50 & 0.60 & 1.60 & 0.30 & 0.20 & 0.20 & 0.15 & 0.15 \\
Optimistic   & 0.70 & 0.55 & 1.90 & 0.35 & 0.22 & 0.18 & 0.12 & 0.13 \\
\bottomrule
\end{tabular}
\end{adjustbox}
\end{table}

Each rubric combines five normalised factors representing simple payback attractiveness, grant support, export-tariff attractiveness, financing attractiveness, and electricity price pressure. Let \(x_j\) denote the normalised score for factor \(j \in \{1,\ldots,5\}\), where scores range from 0 to 100 and 50 represents a neutral condition. For behavioural regime \(r\), the rubric-specific weights \(w_{r,j}\) satisfy \(\sum_{j=1}^{5} w_{r,j}=1\). The unbounded behavioural score is:

\begin{equation}
\tilde{B}_r = 50 + \sum_{j=1}^{5} w_{r,j}(x_j - 50)
\label{eq:behavioural_score_raw}
\end{equation}

and is clipped to the admissible range:

\begin{equation}
B_r = \min\left(100, \max\left(0, \tilde{B}_r\right)\right)
\label{eq:behavioural_score_clipped}
\end{equation}

The clipped score is converted into a multiplicative adjustment factor:

\begin{equation}
\tilde{m}_r = 1 + s_r \left(\frac{B_r - 50}{50}\right)
\label{eq:behavioural_multiplier_raw}
\end{equation}
where \(s_r\) is the sensitivity parameter. Thus, \(B_r = 50\) gives \(\tilde{m}_r = 1\), leaving the calibrated adoption probability unchanged. Values above 50 increase adoption propensity, while values below 50 reduce it. To avoid unrealistic amplification or suppression, the multiplier is bounded by regime-specific limits:

\begin{equation}
m_r = \min\left(m^{\max}_r, \max\left(m^{\min}_r, \tilde{m}_r\right)\right)
\label{eq:behavioural_multiplier_clipped}
\end{equation}

The adjusted adoption probability is then:

\begin{equation}
p_{\text{adj}} = \min\left(1, \max\left(0, p_{\text{base}} m_r\right)\right)
\label{eq:adjusted_probability}
\end{equation}

The final adoption outcome is generated probabilistically using \(p_{\text{adj}}\), preserving the stochastic structure of the baseline ABM. Three behavioural regimes are considered: conservative, balanced, and optimistic, representing modest, central, and stronger bounded behavioural responsiveness, respectively.

\subsection{Structured Scenario Specification}
\label{sec:scenario_specification}

The second augmentation layer concerns structured techno-economic scenario specifications. These scenarios are used for exploratory policy analysis rather than deterministic forecasting. Each scenario modifies only existing ABM inputs, including grant support, FiT level, loan rate, electricity price growth, PV cost trajectory, and export-related assumptions. No new model variables are introduced.

The scenario set includes the calibrated baseline and six exploratory techno-economic conditions, described in Section~\ref{sec:scenario_experiments}.

Before simulation, each scenario is validated using rule-based checks. The validation procedure verifies that required fields are present, numerical ranges are valid, lower bounds do not exceed upper bounds, values remain within admissible policy limits, and uncertainty ranges are not degenerate unless intentionally fixed. In the final experiment set, all exploratory scenarios passed validation, while the calibrated baseline generated expected warnings because selected policy parameters were fixed rather than uncertain.

\subsection{Evaluation Metrics}
\label{sec:evaluation_metrics}

The main outcome variables are cumulative PV adoption by 2040, cumulative public cost by 2040, and public cost per adopter. For each experiment configuration, endpoint adoption and cost are summarised using the mean and standard deviation across Monte Carlo worlds. Public cost per adopter is computed as:

\begin{equation}
C_{\text{adopter}} =
\frac{\bar{C}_{2040}}{\bar{A}_{2040}}
\label{eq:cost_per_adopter}
\end{equation}
where \(\bar{C}_{2040}\) and \(\bar{A}_{2040}\) denote mean cumulative public cost and mean cumulative adoption by 2040 across repeated stochastic evaluations.

Behavioural uplift is calculated relative to the corresponding logistic baseline under the same policy and scenario setting:

\begin{equation}
U_r =
\frac{A^{r}_{2040} - A^{\text{logistic}}_{2040}}
{A^{\text{logistic}}_{2040}}
\times 100
\label{eq:behavioural_uplift}
\end{equation}
where \(A^{r}_{2040}\) denotes cumulative adoption under the specified behavioural regime \(r \in \{\text{conservative}, \text{balanced}, \text{optimistic}\}\). The corresponding relative public cost increase is:

\begin{equation}
\Delta C_r =
\frac{C^{r}_{2040} - C^{\text{logistic}}_{2040}}
{C^{\text{logistic}}_{2040}}
\times 100
\label{eq:cost_increase}
\end{equation}
where \(C^{r}_{2040}\) denotes cumulative public cost under regime \(r\). Scenario-level adoption and cost changes are computed relative to the calibrated baseline scenario under the same behavioural regime, preventing scenario effects from being confounded with behavioural-regime effects.

\section{Experimental Design}
\label{sec:experimental_design}

The experiments evaluate whether the proposed augmentation framework remains stable, interpretable, and policy-relevant when integrated into the calibrated ABM. The design varies three dimensions: policy setting, behavioural regime, and techno-economic scenario.

\subsection{Policy and Behavioural Settings}
\label{sec:policy_behavioural_settings}

Three baseline policy settings represent increasing levels of public support: 40\% grant, 3.00\% APR loan, and 0.15 EUR/kWh FiT; 50\% grant, 5.00\% APR loan, and 0.18 EUR/kWh FiT; and 60\% grant, 6.40\% APR loan, and 0.20 EUR/kWh FiT. For each policy, four behavioural modes are evaluated: the original calibrated logistic model, conservative augmentation, balanced augmentation, and optimistic augmentation. The behavioural experiments test whether adoption follows the expected ordering:

\begin{equation}
A_{\text{logistic}} \leq A_{\text{conservative}} \leq A_{\text{balanced}} \leq A_{\text{optimistic}}
\label{eq:experimental_monotonicity}
\end{equation}

Behavioural diagnostics include the mean behavioural score, mean multiplier, minimum multiplier, and maximum multiplier for each policy and rubric.

\subsection{Scenario Experiments}
\label{sec:scenario_experiments}

The scenario experiments evaluate the calibrated baseline and six exploratory techno-economic scenarios: energy crisis, rapid technology improvement, subsidy withdrawal, green transition acceleration, weak export incentives, and financial tightening. Each scenario is evaluated under the same behavioural modes to examine whether behavioural augmentation differs under supportive, adverse, and mixed policy-market conditions.

For each scenario and behavioural mode, the model reports cumulative adoption by 2040, cumulative public cost by 2040, public cost per adopter, and adoption and cost changes relative to the calibrated baseline scenario under the same behavioural mode.

\subsection{Robustness Evaluation}
\label{sec:robustness_evaluation}

The final robustness evaluation uses \(N_{\text{MC}} = 500\) Monte Carlo worlds and five random seeds. Common random numbers are used where appropriate to support fair comparisons across model configurations. For each configuration, the evaluation records mean endpoint adoption, standard deviation, 95\% confidence intervals, cumulative public cost, cost per adopter, coefficient of variation, and seed-to-seed variability.

Robustness is assessed using four criteria: monotonic behavioural ordering across policies, scenarios, and seeds; bounded behavioural multipliers without saturation; stable scenario rankings across seeds; and absence of pathological variance or excessive seed sensitivity. These checks evaluate whether the LLM-assisted augmentation layers add behavioural and scenario flexibility without destabilising the calibrated ABM.

\section{Results}
\label{sec:results}

\subsection{Baseline Policy Outcomes}
\label{sec:baseline_results}

The calibrated logistic baseline establishes the reference outcome for the three policy settings considered in the experiments. Stronger policy support increases adoption but also raises public expenditure. Adoption increases from 3,094.4 farms under the lowest-support policy to 3,478.6 farms under the highest-support policy, while public cost rises from EUR~16.19 million to EUR~24.82 million. Cost per adopter also increases from EUR~5,231 to EUR~7,135, indicating that higher adoption is achieved with reduced cost-effectiveness.

\subsection{Effect of Behavioural Augmentation}
\label{sec:behavioural_results}

Table~\ref{tab:behavioural_results} shows that behavioural augmentation produces a consistent monotonic ordering across all policy settings. The conservative rubric produces adoption increases of approximately 4.7\%, the balanced rubric increases adoption by 6.2--6.5\%, and the optimistic rubric increases adoption by 11.2--12.8\% relative to the corresponding logistic baseline. The largest policy-level behavioural uplift occurs under the highest-support policy, where optimistic augmentation increases adoption from 3,478.6 to 3,924.4 farms. Public cost increases similarly, reaching a maximum increase of 15.09\% under the same policy. The maximum observed multiplier in this policy comparison is 1.237, indicating bounded behavioural adjustment.

\begin{table}[htbp]
\centering
\caption{Behavioural augmentation results across baseline policy settings.}
\label{tab:behavioural_results}
\setlength{\tabcolsep}{7pt}
\renewcommand{\arraystretch}{1.15}
\begin{adjustbox}{max width=\linewidth}
\begin{tabular}{llrrrr}
\toprule
Policy & Behaviour & Adoption & Cost (EUR M) & Adoption uplift & Cost increase \\
\midrule
40\% grant & Logistic     & 3,094.4 & 16.19 & 0.00\%  & 0.00\% \\
40\% grant & Conservative & 3,241.9 & 17.10 & 4.77\%  & 5.65\% \\
40\% grant & Balanced     & 3,294.7 & 17.45 & 6.47\%  & 7.78\% \\
40\% grant & Optimistic   & 3,442.4 & 18.36 & 11.25\% & 13.43\% \\
\midrule
50\% grant & Logistic     & 3,293.2 & 20.55 & 0.00\%  & 0.00\% \\
50\% grant & Conservative & 3,447.5 & 21.68 & 4.69\%  & 5.52\% \\
50\% grant & Balanced     & 3,497.6 & 22.08 & 6.21\%  & 7.45\% \\
50\% grant & Optimistic   & 3,680.8 & 23.42 & 11.77\% & 13.98\% \\
\midrule
60\% grant & Logistic     & 3,478.6 & 24.82 & 0.00\%  & 0.00\% \\
60\% grant & Conservative & 3,644.6 & 26.21 & 4.77\%  & 5.59\% \\
60\% grant & Balanced     & 3,704.1 & 26.74 & 6.48\%  & 7.72\% \\
60\% grant & Optimistic   & 3,924.4 & 28.57 & 12.81\% & 15.09\% \\
\bottomrule
\end{tabular}
\end{adjustbox}
\end{table}

\begin{figure}[htbp]
\centering
\includegraphics[width=0.9\linewidth]{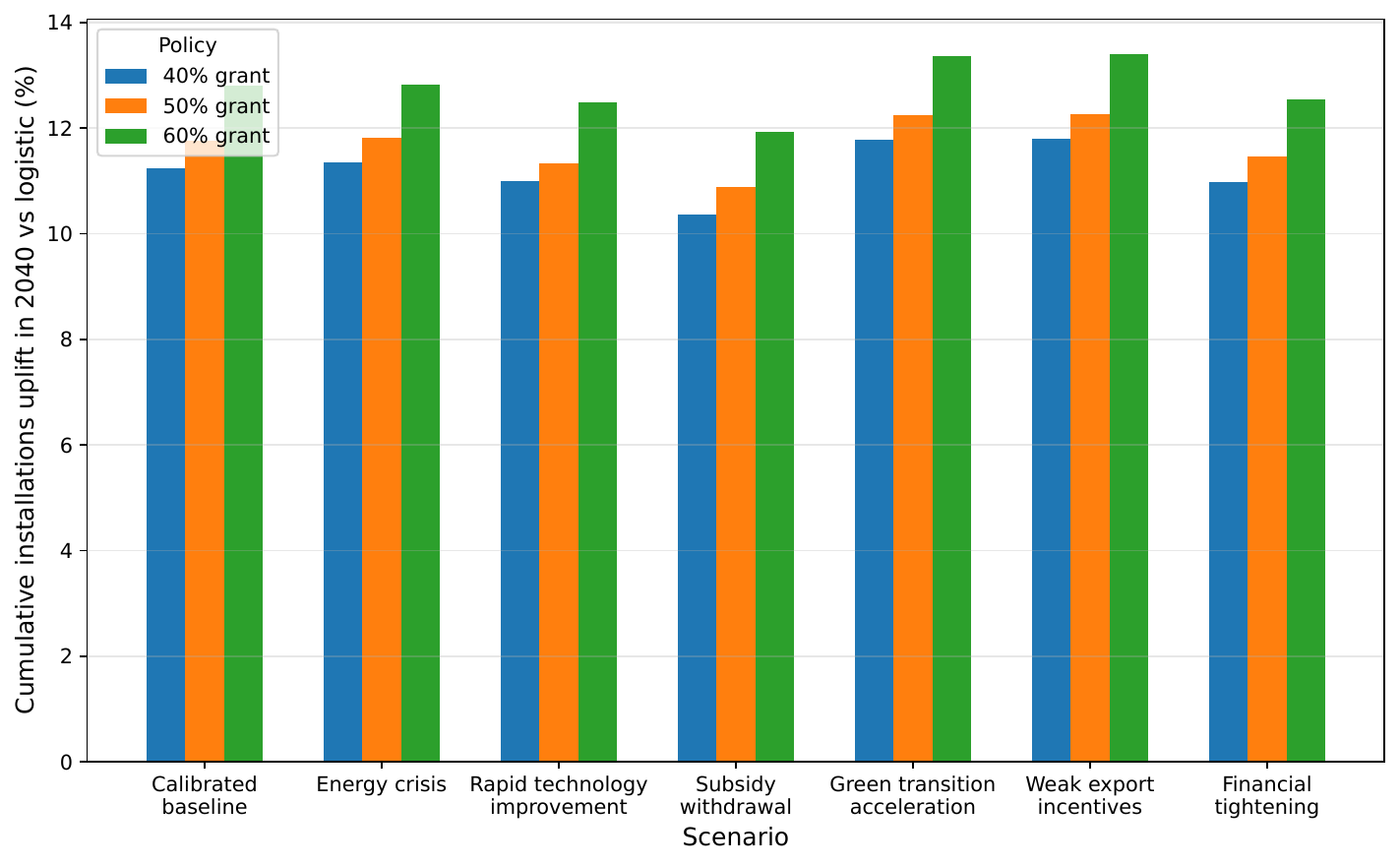}
\caption{Behavioural adoption increase under the optimistic rubric relative to the corresponding logistic case. Bars report percentage increases in cumulative installations in 2040 across scenario and policy settings.}
\label{fig:behavioural_uplift}
\end{figure}

Figure~\ref{fig:behavioural_uplift} shows that optimistic behavioural uplift varies across scenarios and policy environments. The largest uplift occurs under the weak export incentives scenario with the highest-support policy, where optimistic behavioural augmentation increases adoption by 13.40\% relative to the corresponding logistic case.

\subsection{Scenario-Based Adoption and Cost Outcomes}
\label{sec:scenario_results}

Table~\ref{tab:scenario_results} summarises scenario outcomes under the balanced behavioural rubric. The energy crisis scenario produces the largest adoption increase, raising adoption by 11.54\% relative to the calibrated baseline. Green transition acceleration also increases adoption, but with a larger cost increase of 13.71\%. In contrast, subsidy withdrawal reduces adoption by 7.73\% and public cost by 20.50\%. Rapid technology improvement reduces public cost by 12.79\% but does not increase adoption in this experiment, suggesting that lower technology costs alone may be insufficient when other conditions are less favourable.

\begin{table}[htbp]
\centering
\caption{Scenario comparison under the balanced behavioural rubric. Adoption and cost changes are relative to the calibrated baseline under the same behavioural mode.}
\label{tab:scenario_results}
\begin{adjustbox}{max width=\linewidth}
\begin{tabular}{lrrrr}
\toprule
Scenario & Adoption & Cost (EUR M) & Adoption change & Cost change \\
\midrule
Calibrated baseline scenario& 3,498.8 & 22.09 & 0.00\% & 0.00\% \\
Energy crisis & 3,902.5 & 23.78 & 11.54\% & 7.68\% \\
Rapid technology improvement & 3,445.0 & 19.26 & -1.54\% & -12.79\% \\
Subsidy withdrawal & 3,228.3 & 17.56 & -7.73\% & -20.50\% \\
Green transition acceleration & 3,803.2 & 25.12 & 8.70\% & 13.71\% \\
Weak export incentives & 3,657.8 & 25.08 & 4.54\% & 13.55\% \\
Financial tightening & 3,556.2 & 21.05 & 1.64\% & -4.69\% \\
\bottomrule
\end{tabular}
\end{adjustbox}
\end{table}

\begin{figure}[htbp]
\centering
\includegraphics[width=0.9\linewidth]{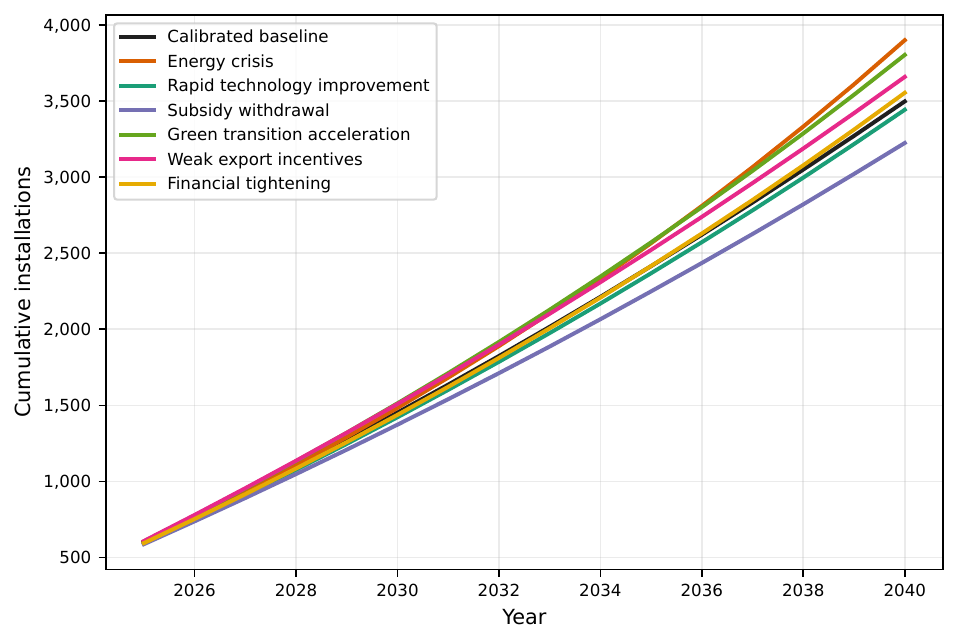}
\caption{Mean cumulative PV adoption trajectories across scenarios under the middle baseline policy and balanced behavioural rubric.}
\label{fig:adoption_trajectories}
\end{figure}

Figure~\ref{fig:adoption_trajectories} shows mean cumulative adoption trajectories across scenarios for the middle policy setting under the balanced behavioural rubric. The energy crisis and green transition acceleration scenarios produce higher adoption pathways, while subsidy withdrawal produces the lowest trajectory.

\begin{figure}[htbp]
\centering
\includegraphics[width=0.9\linewidth]{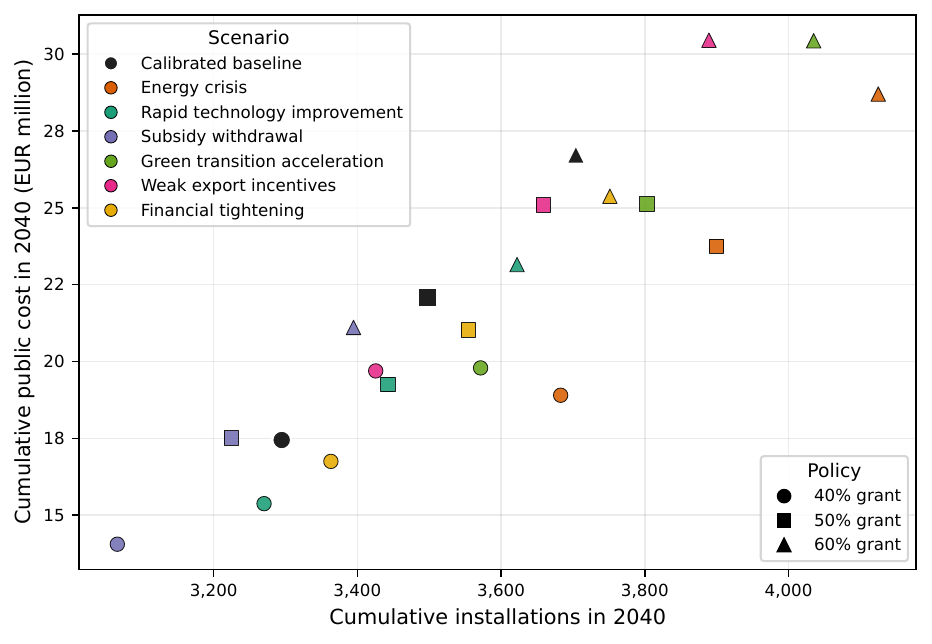}
\caption{Adoption-cost outcomes across scenario and policy settings under the balanced behavioural specification. Points show mean 2040 cumulative installations and cumulative public cost for each scenario-policy combination.}
\label{fig:adoption_cost}
\end{figure}

Figure~\ref{fig:adoption_cost} shows the adoption-cost relationship across scenario and policy settings under the balanced behavioural specification. Higher-adoption combinations generally require higher public expenditure. Rapid technology improvement lowers public cost without producing the highest adoption, while green transition acceleration increases adoption with substantially greater fiscal exposure.

\subsection{Robustness and Stability Analysis}
\label{sec:robustness_results}

Table~\ref{tab:robustness_results} reports the final robustness diagnostics. All scenarios preserve monotonic behavioural ordering, and no scenario produces a saturation or instability flag. Coefficients of variation remain low in the scenario-level summary, ranging from 0.00299 to 0.00409. All six exploratory scenarios pass rule-based validation. The calibrated baseline generates expected warnings because selected policy parameters are fixed rather than uncertain.

In addition to the scenario-level diagnostics in Table~\ref{tab:robustness_results}, the full policy--scenario--behaviour output matrix was inspected for diagnostic flags. Across this full matrix, no scenario fails validation and no seed-sensitive experiment rows are detected. The maximum observed behavioural multiplier is 1.254, confirming bounded behavioural adjustment across the evaluated configurations. Additional multiplier diagnostics are provided in~\ref{app:behavioural_diagnostics}.

\begin{table}[htbp]
\centering
\caption{Robustness diagnostics across scenario specifications. Seed SD denotes seed-to-seed standard deviation, and CV denotes the coefficient of variation.}
\label{tab:robustness_results}
\setlength{\tabcolsep}{6pt}
\renewcommand{\arraystretch}{1.12}
\begin{adjustbox}{max width=\linewidth}
\begin{tabular}{lrrrr}
\toprule
Scenario & Seed SD & CV & Monotonic & Saturation \\
\midrule
Calibrated baseline & 12.61 & 0.00362 & True & False \\
Energy crisis & 15.61 & 0.00403 & True & False \\
Rapid technology improvement & 11.18 & 0.00326 & True & False \\
Subsidy withdrawal & 13.18 & 0.00409 & True & False \\
Green transition acceleration & 13.81 & 0.00366 & True & False \\
Weak export incentives & 13.30 & 0.00367 & True & False \\
Financial tightening & 10.58 & 0.00299 & True & False \\
\bottomrule
\end{tabular}
\end{adjustbox}
\end{table}

\section{Discussion}
\label{sec:discussion}

The results show that LLM-assisted structures can be integrated into a calibrated energy adoption ABM without replacing the underlying techno-economic model. In this framework, the LLM is used offline to support behavioural specification design, while the final behavioural rubrics and scenario specifications are manually inspected, rule-validated, and encoded as deterministic model inputs. The value of the LLM component in this study is therefore methodological rather than predictive: the framework does not claim that LLM-assisted rubrics are superior to expert-designed behavioural rules, but demonstrates how such specifications can be converted into explicit, bounded, and reproducible ABM inputs while preserving model transparency.

Although demonstrated using PV adoption by Irish dairy farms, the integration pattern is more general. The approach can be applied to other calibrated ABMs where model inputs, behavioural factors, and scenario parameters can be explicitly defined and constrained. In such settings, LLM assistance can support specification design, while rule-based validation and fixed deterministic inputs preserve transparency and reproducibility. The key requirement is that LLM outputs are not used as unconstrained agents, but are translated into bounded model inputs before simulation.

The behavioural results indicate coherent and bounded augmentation across policy settings. Adoption follows the expected ordering from the logistic baseline to conservative, balanced, and optimistic regimes, and no saturation behaviour is observed. The strongest behavioural effects occur under more favourable policy and scenario conditions, suggesting that behavioural responsiveness is shaped by the wider techno-economic environment rather than by behavioural assumptions alone.

The scenario results show that PV adoption depends on combinations of policy and market conditions. Energy crisis and green transition acceleration scenarios increase adoption but also raise public cost, while subsidy withdrawal reduces fiscal exposure but suppresses adoption. Rapid technology improvement lowers cost but does not produce the highest adoption, suggesting that lower PV costs alone may be insufficient when other conditions are less favourable.

The robustness results further support the stability of the framework. Behavioural ordering remains stable, no scenario fails validation, coefficients of variation remain low, and no seed-sensitive or saturation behaviour is detected. These diagnostics address a key concern with LLM-assisted simulation: that LLM-assisted components may introduce uninterpretable or unstable behaviour. In this study, the LLM-assisted components remain explicit, bounded, and reproducible.

\section{Conclusion}
\label{sec:conclusion}

This paper proposed a hybrid framework for LLM-assisted specification design in agent-based energy adoption modelling. The novelty of the approach lies in using LLM assistance not as an autonomous simulation agent, but as a controlled mechanism for supporting bounded behavioural rubric construction and structured scenario specification within a calibrated ABM. This preserves the calibrated techno-economic adoption model while allowing richer behavioural and scenario exploration.

The results show stable and interpretable outcomes. Behavioural augmentation increases adoption in the expected order from conservative to balanced and optimistic regimes, while remaining bounded and free from saturation effects. Scenario experiments show that adoption and public cost are shaped by wider policy and market conditions, and robustness checks using 500 Monte Carlo worlds and five seeds confirm stable behavioural ordering, low seed sensitivity, and no pathological variance.

Several limitations remain. The LLM is used offline, so the framework does not evaluate live LLM inference or adaptive LLM agents during simulation. The behavioural rubrics are structured approximations rather than empirical measurements of farmer decision-making, and the scenarios are exploratory rather than predictive. The study also does not compare LLM-assisted rubrics with expert-designed alternatives or alternative prompt outputs, and the specifications may be affected by prompt wording, model bias, and researcher judgement. Future work should incorporate expert or farmer input and test whether LLM-assisted specifications improve empirical behavioural validity.

The study demonstrates that LLM-assisted specification design can provide controlled augmentation layers for calibrated energy ABMs, while preserving transparency, empirical grounding, and reproducibility.

\section*{Acknowledgements}
This publication has emanated from research conducted with the financial support of Research Ireland under Grant number [21/FFP-A/9040].

\appendix
\renewcommand{\thesection}{Appendix~\Alph{section}}

\section{LLM Prompting Summary}
\label{app:llm_prompting}

The LLM-assisted behavioural specifications were generated using ChatGPT-5.5 in an offline prompting workflow. The prompt followed a constrained JSON-generation format, and the LLM was not queried during ABM execution.

For behavioural augmentation, the following prompt was used:

\begin{quote}
You are designing a structured behavioural decision rubric for an agent-based model of solar PV adoption by Irish dairy farms.

The ABM already has a calibrated baseline logistic adoption probability:

\[
p_{\mathrm{base}} = \frac{1}{1 + \exp(-(\alpha \cdot utility + \beta))}
\]
where utility is the NPV-based economic utility of adoption.

Your task is not to replace the calibrated model. Instead, generate a JSON behavioural rubric that adjusts \(p_{\mathrm{base}}\) using interpretable techno-economic and behavioural factors.

Available variables:
payback\_\allowbreak score, grant\_\allowbreak score,
fit\_\allowbreak export\_\allowbreak score,
loan\_\allowbreak score, and price\_\allowbreak score.

Rules:
Return valid JSON only. Include three rubrics: conservative, balanced, optimistic. Each rubric must include sensitivity,
multiplier\_\allowbreak min,
multiplier\_\allowbreak max, and
factor\_\allowbreak weights. The factor weights must include
payback\_\allowbreak score,
grant\_\allowbreak score,
fit\_\allowbreak export\_\allowbreak score,
loan\_\allowbreak score, and
price\_\allowbreak score. Weights must sum to 1.0 for each rubric. Conservative should produce modest behavioural adjustment. Balanced should produce moderate behavioural adjustment. Optimistic should produce stronger but still bounded behavioural adjustment. Avoid unrealistic market saturation. Do not introduce new variables.
\end{quote}

The accepted behavioural rubrics were manually inspected and refined for schema validity, valid weight sums, bounded multipliers, monotonic behavioural ordering, and consistency with the calibrated ABM. The final accepted values are reported in Table~\ref{tab:behavioural_rubrics}.

\section{Additional Behavioural Diagnostics}
\label{app:behavioural_diagnostics}

\begin{figure}[htbp]
\centering
\includegraphics[width=0.9\linewidth]{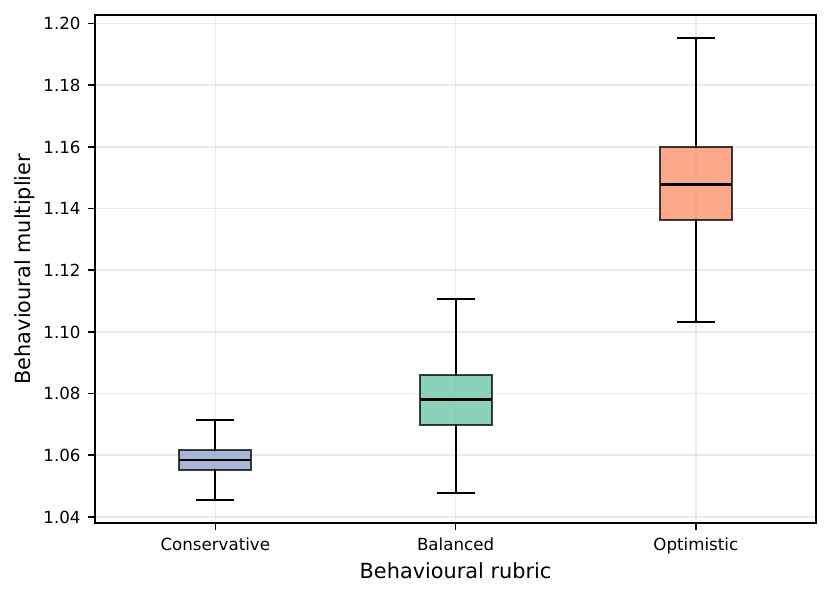}
\caption{Behavioural multiplier diagnostics. Boxplots show the distribution of behavioural multipliers by rubric across Monte Carlo worlds. The distributions indicate that behavioural adjustments remain bounded and ordered across conservative, balanced, and optimistic rubric settings.}
\label{fig:multiplier_distribution_appendix}
\end{figure}

\bibliographystyle{unsrt}  
\bibliography{references}  %%% Remove comment to use the external .bib file (using bibtex).

@incollection{rogers2014diffusion,
  title={Diffusion of innovations},
  author={Rogers, Everett M and Singhal, Arvind and Quinlan, Margaret M},
  booktitle={An integrated approach to communication theory and research},
  pages={432--448},
  year={2014},
  publisher={Routledge}
}

@article{kiesling2012agent,
  title={Agent-based simulation of innovation diffusion: a review},
  author={Kiesling, Elmar and G{\"u}nther, Markus and Stummer, Christian and Wakolbinger, Lea M},
  journal={Central European Journal of Operations Research},
  volume={20},
  number={2},
  pages={183--230},
  year={2012},
  publisher={Springer}
}

@article{rai2015agent,
  title={Agent-based modeling of energy technology adoption: Empirical integration of social, behavioral, economic, and environmental factors},
  author={Rai, Varun and Robinson, Scott A},
  journal={Environmental Modelling \& Software},
  volume={70},
  pages={163--177},
  year={2015},
  publisher={Elsevier}
}

@article{palmer2015modeling,
  title={Modeling the diffusion of residential photovoltaic systems in {Italy}: An agent-based simulation},
  author={Palmer, Johannes and Sorda, Giovanni and Madlener, Reinhard},
  journal={Technological Forecasting and Social Change},
  volume={99},
  pages={106--131},
  year={2015},
  publisher={Elsevier}
}

@article{gao2024large,
  title={Large language models empowered agent-based modeling and simulation: A survey and perspectives},
  author={Gao, Chen and Lan, Xiaochong and Li, Nian and Yuan, Yuan and Ding, Jingtao and Zhou, Zhilun and Xu, Fengli and Li, Yong},
  journal={Humanities and Social Sciences Communications},
  volume={11},
  number={1},
  pages={1--24},
  year={2024},
  publisher={Palgrave}
}

@inproceedings{gurcan2024llm,
  title={Llm-augmented agent-based modelling for social simulations: Challenges and opportunities},
  author={G{\"u}rcan, {\"O}nder},
  booktitle={HHAI 2024: Hybrid Human AI Systems for the Social Good: Proceedings of the Third International Conference on Hybrid Human-Artificial Intelligence},
  pages={134--144},
  year={2024},
  organization={SAGE Publications 1 Oliver's Yard, 55 City Road, London, EC1Y 1SP}
}

@inproceedings{park2023generative,
  title={Generative agents: Interactive simulacra of human behavior},
  author={Park, Joon Sung and O'Brien, Joseph and Cai, Carrie Jun and Morris, Meredith Ringel and Liang, Percy and Bernstein, Michael S},
  booktitle={Proceedings of the 36th annual acm symposium on user interface software and technology},
  pages={1--22},
  year={2023}
}

@article{ji2023survey,
  title={Survey of hallucination in natural language generation},
  author={Ji, Ziwei and Lee, Nayeon and Frieske, Rita and Yu, Tiezheng and Su, Dan and Xu, Yan and Ishii, Etsuko and Bang, Ye Jin and Madotto, Andrea and Fung, Pascale},
  journal={ACM computing surveys},
  volume={55},
  number={12},
  pages={1--38},
  year={2023},
  publisher={ACM New York, NY}
}

@article{larooij2025large,
  title={Do large language models solve the problems of agent-based modeling},
  author={Larooij, Maik and T{\"o}rnberg, Petter},
  journal={A Critical Review of Generative Social Simulations},
  year={2025}
}

@article{wang2025large,
  title={Large language models that replace human participants can harmfully misportray and flatten identity groups},
  author={Wang, Angelina and Morgenstern, Jamie and Dickerson, John P},
  journal={Nature Machine Intelligence},
  volume={7},
  number={3},
  pages={400--411},
  year={2025},
  publisher={Nature Publishing Group UK London}
}

@article{robinson2015determinants,
  title={Determinants of spatio-temporal patterns of energy technology adoption: An agent-based modeling approach},
  author={Robinson, Scott A and Rai, Varun},
  journal={Applied Energy},
  volume={151},
  pages={273--284},
  year={2015},
  publisher={Elsevier}
}

@article{pearce2018feed,
  title={Feed-in tariffs for solar microgeneration: Policy evaluation and capacity projections using a realistic agent-based model},
  author={Pearce, Phoebe and Slade, Raphael},
  journal={Energy Policy},
  volume={116},
  pages={95--111},
  year={2018},
  publisher={Elsevier}
}

@article{peralta2022spatio,
  title={Spatio-temporal modelling of solar photovoltaic adoption: An integrated neural networks and agent-based modelling approach},
  author={Peralta, Ali Alderete and Balta-Ozkan, Nazmiye and Longhurst, Philip},
  journal={Applied Energy},
  volume={305},
  pages={117949},
  year={2022},
  publisher={Elsevier}
}

@article{zhang2022agent,
  title={An agent-based diffusion model for Residential Photovoltaic deployment in {Singapore}: Perspective of consumers' behaviour},
  author={Zhang, Nan and Lu, Yujie and Chen, Jiayu and Hwang, Bon-Gang},
  journal={Journal of Cleaner Production},
  volume={367},
  pages={132793},
  year={2022},
  publisher={Elsevier}
}

@article{zhang2019empirically,
  title={Empirically grounded agent-based models of innovation diffusion: a critical review},
  author={Zhang, Haifeng and Vorobeychik, Yevgeniy},
  journal={Artificial Intelligence Review},
  volume={52},
  number={1},
  pages={707--741},
  year={2019},
  publisher={Springer}
}

@article{grimm2020odd,
  title={The {ODD} protocol for describing agent-based and other simulation models: A second update to improve clarity, replication, and structural realism},
  author={Grimm, Volker and Railsback, Steven F and Vincenot, Christian E and Berger, Uta and Gallagher, Cara and DeAngelis, Donald L and Edmonds, Bruce and Ge, Jiaqi and Giske, Jarl and Groeneveld, Juergen and others},
  journal={Journal of Artificial Societies and Social Simulation},
  volume={23},
  number={2},
  year={2020}
}

@article{thiele2014facilitating,
  title={Facilitating parameter estimation and sensitivity analysis of agent-based models: A cookbook using {NetLogo} and {R}},
  author={Thiele, Jan C and Kurth, Winfried and Grimm, Volker},
  journal={Journal of Artificial Societies and Social Simulation},
  volume={17},
  number={3},
  pages={11},
  year={2014}
}

@article{troost2023keep,
  title={How to keep it adequate: A protocol for ensuring validity in agent-based simulation},
  author={Troost, Christian and Huber, Robert and Bell, Andrew R and Van Delden, Hedwig and Filatova, Tatiana and Le, Quang Bao and Lippe, Melvin and Niamir, Leila and Polhill, J Gareth and Sun, Zhanli and others},
  journal={Environmental Modelling \& Software},
  volume={159},
  pages={105559},
  year={2023},
  publisher={Elsevier}
}

@article{pfenninger2014energy,
  title={Energy systems modeling for twenty-first century energy challenges},
  author={Pfenninger, Stefan and Hawkes, Adam and Keirstead, James},
  journal={Renewable and sustainable energy reviews},
  volume={33},
  pages={74--86},
  year={2014},
  publisher={Elsevier}
}

@article{yue2018review,
  title={A review of approaches to uncertainty assessment in energy system optimization models},
  author={Yue, Xiufeng and Pye, Steve and DeCarolis, Joseph and Li, Francis GN and Rogan, Fionn and Gallach{\'o}ir, Brian {\'O}},
  journal={Energy strategy reviews},
  volume={21},
  pages={204--217},
  year={2018},
  publisher={Elsevier}
}

@article{binsted2020evaluating,
  title={Evaluating long-term model-based scenarios of the energy system},
  author={Binsted, Matthew and Iyer, Gokul and Cui, Ryna and Khan, Zarrar and Dorheim, Kalyn and Clarke, Leon},
  journal={Energy Strategy Reviews},
  volume={32},
  pages={100551},
  year={2020},
  publisher={Elsevier}
}

@inproceedings{aher2023using,
  title={Using large language models to simulate multiple humans and replicate human subject studies},
  author={Aher, Gati V and Arriaga, Rosa I and Kalai, Adam Tauman},
  booktitle={International conference on machine learning},
  pages={337--371},
  year={2023},
  organization={PMLR}
}

@article{argyle2023out,
  title={Out of one, many: Using language models to simulate human samples},
  author={Argyle, Lisa P and Busby, Ethan C and Fulda, Nancy and Gubler, Joshua R and Rytting, Christopher and Wingate, David},
  journal={Political Analysis},
  volume={31},
  number={3},
  pages={337--351},
  year={2023},
  publisher={Cambridge University Press}
}

@article{faiud2024agent,
  title={An agent-based modeling approach for simulating solar {PV} adoption: A case study of {Irish} dairy farms},
  author={Faiud, Iias and Schukat, Michael and Mason, Karl},
  journal={Renewable Energy Focus},
  volume={51},
  pages={100653},
  year={2024},
  publisher={Elsevier}
}

@inproceedings{faiud2023modelling,
  title={Modelling solar {PV} adoption in {Irish} dairy farms using agent-based modelling},
  author={Faiud, Iias and Mason, Karl and Schukat, Michael},
  booktitle={Joint European Conference on Machine Learning and Knowledge Discovery in Databases},
  pages={292--300},
  year={2023},
  organization={Springer}
}

\end{document}